\documentclass[letterpaper]{article} 
\usepackage{aaai2026}  
\usepackage{times}  
\usepackage{helvet}  
\usepackage{courier}  
\usepackage[hyphens]{url}  
\usepackage{graphicx} 
\usepackage{natbib}  
\usepackage{caption} 
\usepackage{algorithm}
\usepackage{algorithmic}
\usepackage{soul}
\usepackage{amsmath}
\usepackage{amsfonts}
\usepackage{tikz}
\usepackage{booktabs}
\usepackage{multirow}
\usepackage{array}
\usepackage{cleveref}
\usepackage{threeparttable}
\usepackage{newfloat}
\usepackage{listings}
\DeclareCaptionStyle{ruled}{labelfont=normalfont,labelsep=colon,strut=off} 
\floatstyle{ruled}
\newfloat{listing}{tb}{lst}{}
\floatname{listing}{Listing}
\newcommand{\method}{ICAD-LLM}
\title{ICAD-LLM: One-for-All Anomaly Detection via In-Context Learning with Large Language Models}
\author{
    Zhongyuan Wu\textsuperscript{\rm 1, \rm3}, 
    Jingyuan Wang\textsuperscript{\rm 1, \rm2, \rm3, \rm4}\thanks{Corresponding author. Email: jywang@buaa.edu.cn}, 
    Zexuan Cheng\textsuperscript{\rm 1, \rm3}, 
    Yilong Zhou\textsuperscript{\rm 1, \rm3}, 
    Weizhi Wang\textsuperscript{\rm 1, \rm3}, \\
    Juhua Pu\textsuperscript{\rm 1, \rm3}, 
    Chao Li\textsuperscript{\rm 1, \rm3}, 
    Changqing Ma\textsuperscript{\rm 5} 
}
\affiliations{
    \textsuperscript{\rm 1}School of Computer Science and Engineering, Beihang University, Beijing, China\\
    \textsuperscript{\rm 2} School of Economics and Management, Beihang University, Beijing, China \\
    \textsuperscript{\rm 3} MOE Engineering Research Center of Advanced Computer Application Technology, Beihang University, China \\
    \textsuperscript{\rm 4} MIIT Key Laboratory of Data Intelligence and Management, Beihang University, Beijing, China \\
    \textsuperscript{\rm 5}Capinfo Co., Ltd.\\

}

\usepackage{bibentry}

\begin{document}

\maketitle
\begin{abstract}
Anomaly detection (AD) is a fundamental task of critical importance across numerous domains. Current systems increasingly operate in rapidly evolving environments that generate diverse yet interconnected data modalities—such as time series, system logs, and tabular records—as exemplified by modern IT systems. Effective AD methods in such environments must therefore possess two critical capabilities: (1) the ability to handle heterogeneous data formats within a unified framework, allowing the model to process and detect multiple modalities in a consistent manner during anomalous events; (2) a strong generalization ability to quickly adapt to new scenarios without extensive retraining. However, most existing methods fall short of these requirements, as they typically focus on single modalities and lack the flexibility to generalize across domains. To address this gap, we introduce a novel paradigm: In-Context Anomaly Detection (ICAD), where anomalies are defined by their dissimilarity to a relevant reference set of normal samples. Under this paradigm, we propose ICAD-LLM, a unified AD framework leveraging Large Language Models' in-context learning abilities to process heterogeneous data within a single model. Extensive experiments demonstrate that ICAD-LLM achieves competitive performance with task-specific AD methods and exhibits strong generalization to previously unseen tasks, which substantially reduces deployment costs and enables rapid adaptation to new environments. To the best of our knowledge, ICAD-LLM is the first model capable of handling anomaly detection tasks across diverse domains and modalities. The extended version of this paper will be available at https://github.com/nobody384/ICAD-LLM.

\end{abstract}



\section{Introduction}
Anomaly detection (AD) is a critical task with pervasive importance across numerous domains. 
\begin{figure}[htb]
    \centering
    \includegraphics[width=\linewidth]{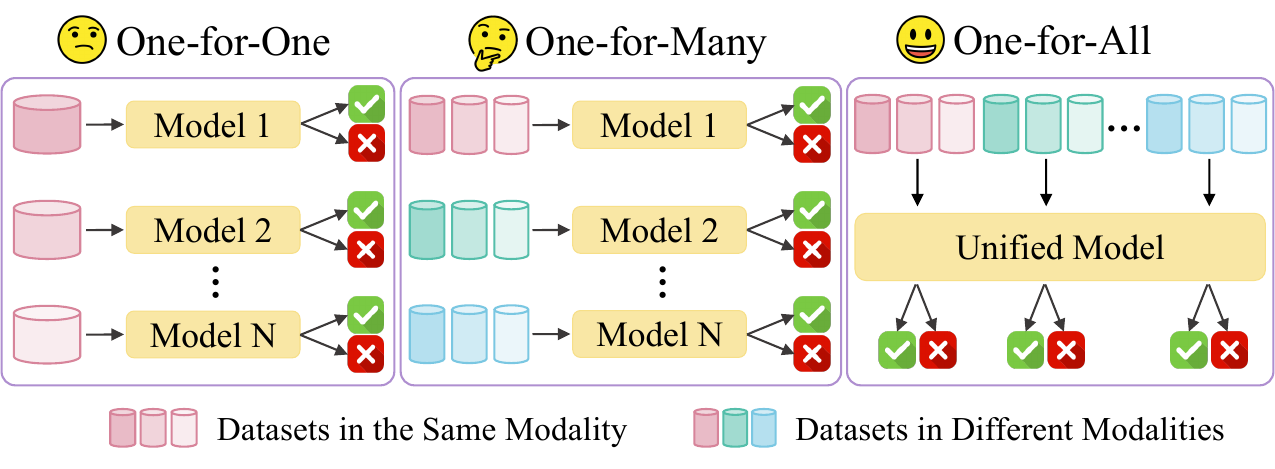}
    \caption{An illustration of different levels of  AD. }
    \label{fig:one-for-all}
\end{figure}
However, most existing AD methods are primarily designed for single data modalities, and their ability to generalize to new, unseen scenarios is often limited \cite{ResAD}. This creates a significant gap between academic research and the demands of real-world applications. For instance, in modern IT systems like e-commerce platforms, a single fault such as a payment failure can manifest concurrently as CPU spikes (time series), error logs (log data), and abnormal transaction records (tabular data). This situation calls for a universal model that can handle various data types effectively, thereby reducing the need for multiple, disparate solutions. Furthermore, as these business systems rapidly evolve with new services and architectures, existing AD models must adapt quickly to novel scenarios without complete retraining for fast deployment. Under these conditions, conventional approaches that require separate model training for each task or modality become operationally infeasible.

In terms of their model-to-task mapping, as shown in \Cref{fig:one-for-all}, current AD methods can be categorized into different levels. Traditional One-for-One AD (OFO-AD) methods train dedicated models for each specific dataset, learning the unique distribution of ``normal'' instances within that particular task \cite{ocsvm,Donut,LogBert,DCdetector,MCM}. While effective for their designated tasks, these methods fail to generalize across different tasks or domains, necessitating costly retraining for each new application scenario.
More recently, One-for-Many AD (OFM-AD) approaches have emerged as a response to the limitations of OFO-AD, improving generalization by enabling a single model to detect anomalies across various predefined tasks within the same modality \cite{UniAD,MambaAD,ResAD,ACR}. However, these approaches still fall short of the requirements outlined above: they remain confined to that single data modality and lack the architectural versatility to be applied to other data types.

The aforementioned limitations of existing AD methods lead to a natural yet ambitious question: Is it possible to develop a ``One-for-All'' model capable of handling diverse tasks across multiple data modalities? To answer this question, we must revisit the original definition of anomalies by Grubbs \cite{grubbs1969procedures}, which described anomalies as ``\textit{one that appears to deviate markedly from other members of the sample in which it occurs}''. This principle, however, is contradicted by the design of most existing AD methods. 
By evaluating each sample individually, these methods prevent the model from directly comparing it to ``other members'' at inference time, which forces the model to rely on an internalized, static understanding of normality learned from the training set.
This act of memorization binds the model to a specific task, restricting it from generalizing across tasks and modalities. This insight leads us to a new perspective: what if we could empower the AD model with the fundamental skill of comparison by providing the necessary context on-the-fly, thereby decoupling AD from task-specific distribution learning? To realize this vision, we introduce In-Context Anomaly Detection (ICAD). ICAD explicitly provides a reference set of normal samples during inference, and then assesses anomalies by comparing the target sample to this contextually relevant reference set. By shifting the objective from memorization to in-context comparison, this approach is inherently more flexible and readily applicable across diverse data modalities.

However, translating this high-level ICAD paradigm into a practical and effective One-for-All model is non-trivial. It necessitates a framework design that satisfies three key requirements (REQ): \textit{\textbf{REQ1}-Feature Alignment}. Given that data from different modalities have vastly different feature dimensions and semantic structures, the model must first project these disparate inputs into a common embedding space, which is the foundational step that enables a single, unified architecture to process them meaningfully. \textit{\textbf{REQ2}-Discrepancy-Sensitive Representation}. The model must extract rich, semantic representations that are not only modality-agnostic but also sensitive to the subtle dissimilarities between a target sample and its reference sets. \textit{\textbf{REQ3}-Task-Agnostic Discriminative Objective}. Unlike traditional AD training objectives (e.g., minimizing reconstruction loss) that are tightly coupled to specific data distributions, ICAD requires a new training objective. This objective must decouple the model from the training data by explicitly training its universal ability to discriminate a target's dissimilarity against its reference set, rather than encouraging task-specific memorization.

To fulfill these requirements, we propose \method{}, a unified AD framework leveraging Large Language Models' in-context learning abilities to process heterogeneous data within a single model. \method{} consists of three key components, each tailored to satisfy a specific requirement. To meet \textit{\textbf{REQ1}}, we design a \textit{Modality-Aware Encoder} that projects heterogeneous inputs from time-series, logs, and tables into a unified, fixed-dimension embedding space. To solve \textit{\textbf{REQ2}}, we employ a \textit{Prompt-Guided Representation Module}. This component harnesses the in-context learning capacity of Large Language Models to extract semantically rich representations. To address \textit{\textbf{REQ3}}, we formulate a \textit{Contextual Contrastive Learning} objective, which explicitly trains the model to discern subtle differences between normal and anomalous patterns. 
Crucially, \method{} is only trained once to acquire a general-purpose anomaly discrimination capability. At inference time, this single model can tackle anomaly detection tasks across diverse modalities without task-specific retraining. This ``train-once, apply-broadly'' strategy equips \method{} with much flexibility and efficiency. 
Extensive experiments show that \method{} achieves performance competitive with state-of-the-art task-specific methods and exhibits strong generalization to previously unseen tasks. 
To the best of our knowledge, \method{} is the first model capable of handling AD tasks across diverse domains and modalities. The main contributions of this paper are summarized as follows:
\begin{itemize}
    \item We introduce In-Context Anomaly Detection, which redefines anomaly detection based on the concept of contextual dissimilarity, enabling a more generalized and flexible anomaly detection approach.
    \item We propose \method{}, a novel model designed to effectively implement the ICAD paradigm across multiple data modalities and diverse tasks.
    \item We demonstrate the \method{} achieves competitive performance on standard AD benchmarks and, more importantly, exhibits strong generalization to out-of-domain datasets without task-specific retraining. 
\end{itemize}
\begin{figure*}[htbp]
    \centering
\includegraphics[width=0.95\textwidth]{figure/paradigm_compare.pdf}
    \caption{Comparison of different AD paradigms. (a) Traditional One-for-One/Many AD learns a fixed decision boundary from training data to identify outliers. (b) In-Context Anomaly Detection learns a general comparison function, detecting anomalies based on their discrepancy from a given reference set.}
    \label{fig:paradigm}
\end{figure*}

\section{Related Work}

\subsection{One-for-One/Many Anomaly Detection}
Traditional anomaly detection often follows a one-for-one paradigm, training a separate model for each dataset. Early approaches include classic machine learning methods~\cite{lof,ocsvm,iforest}, while recent deep learning models adopt boundary-based formulations that learn compact hyperspheres around normal data. Representative works include OmniAnomaly~\cite{OmniAnomaly}, AnomalyTransformer~\cite{AnomalyTrans}, and DCdetector~\cite{DCdetector} for time series; MCM~\cite{MCM} for tabular data; and LogAnomaly~\cite{LogAnomaly} and LogBert~\cite{LogBert} for system logs. Although effective in their respective domains, these models degrade when faced with unseen tasks. The one-for-many paradigm addresses this by building a single model for multiple datasets within the same modality. Solutions include unified architectures such as PatchCore~\cite{PatchCore}, SimpleNet~\cite{SimpleNet}, GOAD~\cite{GOAD}, UniAD~\cite{UniAD}, and MambaAD~\cite{MambaAD}, and adaptations of large-scale pre-trained models, e.g., PMAD~\cite{PMAD}, WinCLIP~\cite{WinCLIP}, ResAD~\cite{ResAD}, and AnomalyLLM~\cite{AnomalyLLM}. Notably, PMAD was the first to explicitly advocate the ``one-for-all'' concept. While these approaches improve intra-modality generalization, they remain restricted to specific data types, limiting their applicability across heterogeneous modalities.

\subsection{Representation Learning with LLMs and Unified Architectures}
Recent studies leverage Large Language Models (LLMs) and unified architectures to learn transferable representations across heterogeneous data. 
LLM-based approaches enrich semantic encoding and contextual understanding~\cite{cheng2025poi,li2023web}, and introduce efficient multi-task adaptation via contextual attention modulation~\cite{pan2025contextual}. 
Unified temporal–spatial frameworks align multi-source sequences in shared embedding spaces~\cite{BIGCITY,han2025bridging}, while self-supervised and generative pretraining~\cite{jiang2023self,ji2023stssl,ren2022gan,ren2021rapt} enhance representation robustness under noise and data scarcity. 
Siamese and graph-based architectures~\cite{ren2022rsd,zhang2024vecCity} further generalize structured or sparse entities, with robustness-oriented designs~\cite{ji2025seeing,liu2024bayesnn} supporting adaptation to unseen tasks. 
Together, these directions establish modality‑agnostic encoding, feature alignment, and resilient representation learning, which are principles underlying unified anomaly detection.


\section{Preliminary}
Let $\mathbb{M} = \left\{\mathcal{M}_i \right\}_{i=1}^m$ be the set of all possible $m$ modalities, and $\mathbb{T}=\bigcup_{i=1}^m T_{\mathcal{M}_i}$ denote the universe of tasks, where $T_{\mathcal{M}_i} = \{\tau_j^{(i)}\}_{j=1}^{n_i}$ represents the set of tasks under modality $\mathcal{M}_i$. For any task $\tau \in \mathbb{T}$, we define the task-specific data as $
    \mathcal{D}_{\tau} = \mathcal{X}_\tau \times \mathcal{Y}_\tau$, 
where $\mathcal{X}_{\tau}$ presents all the samples in task $\tau$, and $\mathcal{Y}_\tau=\{0,1\}$ denotes the binary labels indicating normal (0) versus anomalous (1) classes. For clarity, we denote the general anomaly detection process as $\mathcal{A}$. \Cref{fig:paradigm} provides a visual comparison of different AD paradigms.

\subsection{One-for-One/Many AD}Given a target sample $x_{tgt} \in \mathcal{X}_\tau$, both OFO/OFM-AD aim to learn a score function $f$, formalizing the AD process as: 
\begin{equation}
\mathcal{A}(x_{tgt}; f, \theta_{\tau}) = \mathbb{I}(f(x_{tgt}) \geq \theta_{\tau}),
\end{equation}
where $\theta_\tau$ is the decision threshold of task $\tau$ and $\mathbb{I}(\cdot)$ is the indicator function. The key distinction lies in their training scope. 
OFO-AD learns $f$ from a task-specific dataset $\mathcal{D}_{train} \subset \mathcal{D}_{\tau}$, specializing in individual tasks. In contrast, for OFM-AD, the function is trained on a composite dataset that comprises multiple tasks within a single modality, i.e., $\mathcal{D}_{train} \subset \bigcup_{\tau \in T_{\mathcal{M}}} D_{\tau}$, where $T_{\mathcal{M}}$ is the set of tasks under modality $\mathcal{M}$.

\subsection{In-Context AD}
Our proposed ICAD paradigm redefines anomaly detection by leveraging contextual comparison. Let $\mathbb{T}_{train} \subset \mathbb{T}$ represent the subset of tasks observed during training, which can encompass a mixture of modalities. For any task $\tau^* \in \mathbb{T}$ (including unseen tasks where $\tau^* \notin \mathbb{T}_{train}$), we define a  reference set $\mathcal{R} = \left\{ r_1, r_2, \cdots, r_K\right\} \subset \left\{x \in \mathcal{X}_{\tau^*}:y=0\right\}$, consisting of $K$ normal samples that characterize the expected behavior for that task.
Given a target sample $x_{tgt} \in \mathcal{X}_{\tau^*}$ and its reference set $\mathcal{R}$, ICAD determines anomalies by computing their contextual discrepancy, and the anomaly detection process can be described as:
\begin{equation}
\mathcal{A}(\mathcal{R},x_{tgt}; \delta, \theta_{\tau}^{\prime}) = \mathbb{I}(\delta(\mathcal{R}, x_{tgt}) \geq \theta_{\tau}^{\prime}),
\label{icad}
\end{equation}
where $\delta(\mathcal{R}, x_{tgt})$ measures the dissimilarity between the target and reference samples, and $\theta_{\tau}^{\prime}$ is the task-specific discrepancy threshold.
By defining anomalies through dynamic comparisons to the provided reference sets, this paradigm decouples the model from a static definition of normality, enabling flexible AD across diverse tasks and modalities.

\begin{figure*}[htbp]
    \centering
    \includegraphics[width=0.85\textwidth]{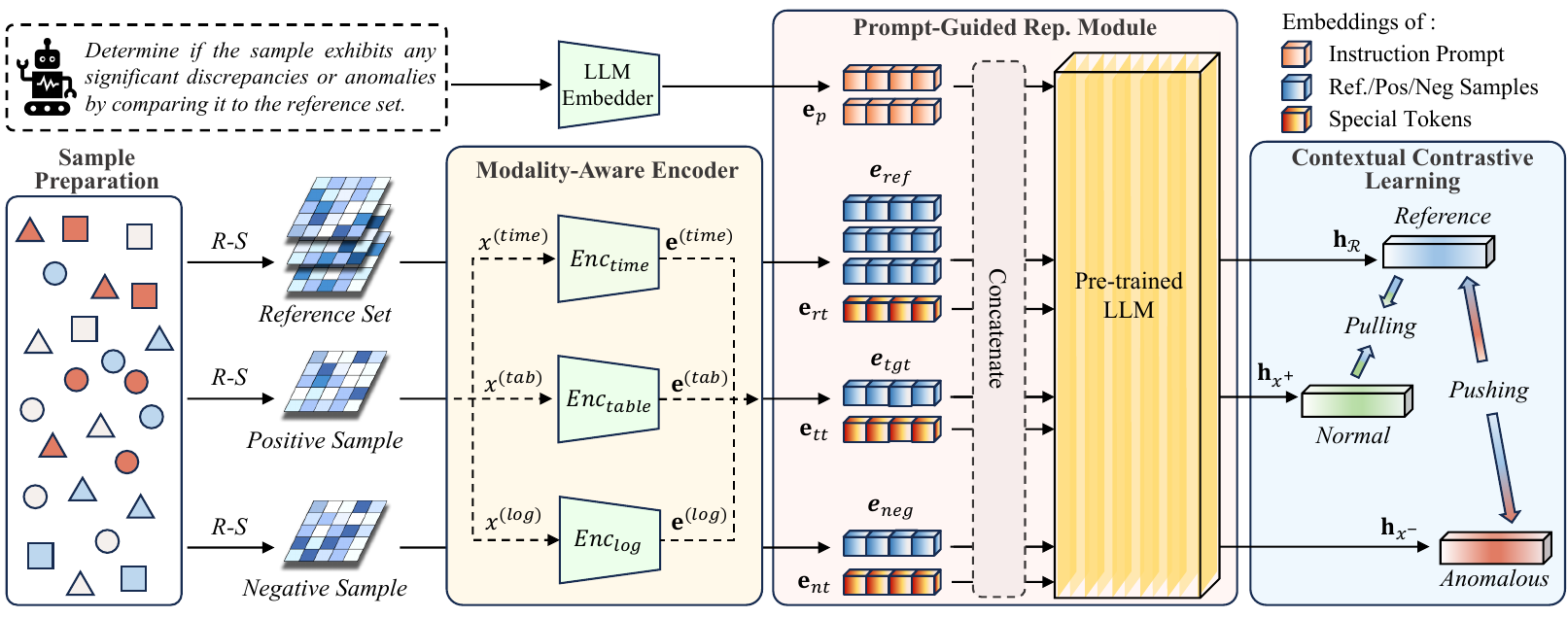}
    \caption{The overall pipeline of \method{}. A reference set and target samples are first encoded by the Modality-Aware Encoder. The resulting embeddings are then concatenated with prompt and special tokens, and fed into the Prompt-Guided Representation Module to extract context-sensitive representations. Finally, the Contextual Contrastive Learning objective is employed to optimize the model's ability to distinguish between these samples. \textit{R-S} means random selection.}
    \label{fig:mydiagram}
\end{figure*}
\section{Methodology}
In this section, we propose \method{}, a
unified AD framework that harnesses the powerful in-context learning abilities of Large Language Models to detect anomalies across multiple data modalities. \Cref{fig:mydiagram} illustrates the overall pipeline of \method{}, which consists of three key components. First, 
a \textit{Modality-Aware Encoder} addresses feature alignment by projecting heterogeneous inputs into a unified embedding space. Second, the \textit{Prompt-Guided Representation Module} uses an LLM to extract modality-agnostic representations that are highly sensitive to subtle dissimilarities. Third, the model is trained with a \textit{Contextual Contrastive Learning (CCL)} objective, which sharpens its discriminative power by maximizing the discrepancy for anomalous samples while minimizing it for normal ones. The resulting discrepancy score is then used for final anomaly detection.

\subsection{Sample Preparation}
Before detailing the model architecture, we first describe how raw, heterogeneous data is transformed into a standardized ``sample'' format, which is the fundamental unit that our model can process.
\subsubsection{Time Series Processing} 
Let $X_{\text{raw}}^{(\text{time})} \in \mathbb{R}^{L \times d_\text{raw}}$ denote a raw time series, where $L$ is the sequence length and $d_{\text{raw}}$ is the feature dimension. We define a patching function that segments $X_{\text{raw}}^{(\text{time})}$ into a set of $n$ patches:
\begin{equation}
\left\{x_i^{(\text{time})}\right\}_{i=1}^n = \text{Patch}(X_{\text{raw}}^{(\text{time})}).
\end{equation}
Each patch $x_i^{(\text{time})} \in \mathbb{R}^{p \times d_\text{raw}}$ is treated as an individual sample, where $p$ is the patch length.

\subsubsection{Tabular Row Processing}
Let $X_{\text{raw}}^{(\text{tab})} = \{x_{\text{raw},i}^{(\text{tab})}\}_{i=1}^{N}$ be a tabular dataset where each row $x_{\text{raw},i}^{(\text{tab})} \in \mathbb{R}^{F_i}$ has $F_i$ features. To ensure uniform dimensionality, we define a padding-truncation function $\Psi(\cdot)$ that maps each row to a fixed dimension $F^{\prime}$ with zero-padding or truncation:
\begin{equation}
x_i^{(\text{tab})} = \Psi(x_{\text{raw},i}^{(\text{tab})}) = 
\begin{cases}
[x_{\text{raw},i}^{(\text{tab})}, \mathbf{0}_{F^{\prime}-F_i}] & \text{if } F_i < F^{\prime}  \\
x_{\text{raw},i}^{(\text{tab})}[0:F^{\prime}] & \text{if } F_i \geq F^{\prime}
\end{cases},
\end{equation}
where $\mathbf{0}_{F^{\prime}-F_i} \in \mathbb{R}^{F^{\prime}-F_i}$ is a zero vector of length $F^{\prime}-F_i$, and $x[0:F^{\prime}]$ denotes selecting the first $F^{\prime}$ elements. Each $x_{\text{raw},i}^{(\text{tab})}$ constitutes a single sample.

\subsubsection{Log Sequence Processing}
Raw log messages $X_{\text{raw}}^{(\text{log})} = \{m_t\}_{t=1}^T$ undergo two-stage processing. First, a log parser\footnote{We use Drain3 (https://github.com/logpai/Drain3) as the log parser.} extracts templates: $S = \{k_t\}_{t=1}^T = \text{Parse}(X_{\text{raw}}^{(\text{log})})$, where $k_t \in \mathcal{K}$ represents the log template at time $t$. Second, temporal windowing partitions $S$ into fixed-size segments: $x_i^{(\text{log})} = S[(i-1)w+1:iw]$, where each window of $w$ consecutive log keys forms a sample.

\subsection{Modality-Aware Encoder}To effectively process heterogeneous samples from various modalities, we employ the Modality-Aware Encoder, which transforms prepared samples from different modalities into a unified embedding space. Given a sample $x^{(\mathcal{M})}$ from modality $\mathcal{M}$, the encoder applies a transformation as:
\begin{equation}
\mathbf{e}^{(\mathcal{M})} = E_{\mathcal{M}}(x^{(\mathcal{M})}),
\end{equation}
where $\mathbf{e}^{(\mathcal{M})} \in \mathbb{R}^{N^{(\mathcal{M})} \times d_{\text{model}}}$ is the encoded embedding, $N^{(\mathcal{M})}$ is the sequence length, and $d_{\text{model}}$ is the embedding dimension.
The specific implementations of $E_{\mathcal{M}}$ are as follows:
For a time series sample $x^{(\text{time})}$, we first apply instance normalization and then feed the result into a Convolutional Neural Network (CNN) to align the feature dimension:
    \begin{equation}
    E_{\text{time}}\left(x^{(\text{time})}\right) = \text{CNN}\left(\text{IN}\left(x^{(\text{time})}\right)\right).
    \end{equation}
For a tabular sample $x^{(\text{tab})}$, following the encoding approach of MCM \cite{MCM}, the encoder $E_{\text{tab}}$ utilizes a two-layer Multilayer Perceptron (MLP) to produce its embedding:
    \begin{equation}
 E_{\text{tab}}\left(x^{(\text{tab})}\right) = \text{MLP}^{(\text{tab})}\left(x^{(\text{tab})}\right).
    \end{equation}
For a log sample $x^{(\text{log})}$, the encoder $E_{\text{log}}$ first uses the LLM's native tokenizer and embedder $\text{Emb}(\cdot)$ to get initial embeddings, which are then refined by a Transformer encoder:
    \begin{equation}
 E_{\text{log}}\left(x^{(\text{log})}\right) = \text{TransEnc}\left(\text{Emb}\left(x^{(\text{log})}\right)\right).
 \end{equation}
%

\subsection{Prompt-Guided Representation Module}
At the core of our ICAD model is the Prompt-Guided Representation Module. It leverages a pre-trained Large Language Model (LLM) as the backbone, harnessing its powerful capabilities for contextual reasoning and semantic understanding to produce modality-agnostic representations that capture the subtle differences between a target sample and its reference set. This module incorporates two key mechanisms: \textit{(1) Instruction-based Priming.} Inspired by prior work \cite{BIGCITY}, we prepend an instruction prompt to the input sequence which explicitly primes the LLM, directing its powerful reasoning abilities towards the specific goal of assessing contextual dissimilarity, rather than general language understanding. \textit{(2) Token-anchored Representation Pooling.} To obtain distinct and high-level representations for both the context and the target, we introduce two special, learnable tokens: [REF\_TOK] and [TGT\_TOK]. These tokens are inserted into the input sequence, compelling the LLM to aggregate and summarize the information of the reference set and the target sample into their respective token positions.

Formally, let $\mathbf{e}_{p} \in \mathbb{R} ^{L \times d_{\text{model}}}$, where $L$ is the sequence length of prompt tokens, be the embedding of the instruction prompt, $\mathbf{e}_{rt} \in \mathbb{R} ^{1 \times d_{\text{model}}} $ and $\mathbf{e}_{tt} \in \mathbb{R} ^{1 \times d_{\text{model}}}$ be the corresponding embeddings for the aforementioned [REF\_TOK] and [TGT\_TOK]. 
For a given reference set $\mathcal{R}$ and a target sample $x$, we denote their embeddings produced by the Modality-Aware Encoder as $\mathbf{e}_{ref} \in \mathbb{R}^{N \times d_{\text{model}}} $ and $\mathbf{e}_{tgt} \in \mathbb{R}^{N \times d_{\text{model}}}$, respectively.
The final input sequence $S$ is formulated as:
\begin{equation}
    S = \text{Concat}\left( \mathbf{e}_{p}, \mathbf{e}_{ref}, \mathbf{e}_{rt},\mathbf{e}_{tgt},\mathbf{e}_{tt} \right).
    \label{eq:input}
\end{equation}
This sequence is then fed into our LLM backbone, and we extract the final-layer hidden states corresponding to the positions of our special tokens. This yields a holistic representation for the reference set, $\mathbf{h}_{\mathcal{R}}\in \mathbb{R}^{d_{\text{model}}}$ and a representation for the target sample, $\mathbf{h}_{x}\in \mathbb{R}^{d_{\text{model}}}$.
These representations encapsulate the essential characteristics of their inputs while being sensitive to their contextual differences, providing a modality-agnostic basis for anomaly detection.

\subsection{Contextual Contrastive Learning}
We propose the Contextual Contrastive Learning (CCL) objective, which creates a clear margin for discrimination as required by the ICAD paradigm by 
pulling normal samples closer to the representation of their reference set, while pushing anomalous samples further away.
To implement this, we formulate the training process around sample triplets, each designed to teach the model a specific aspect of contextual comparison. For each training step, given a source dataset $\mathcal{D}$ from modality $\mathcal{M}$,  we construct a triplet $\left(\mathcal{R},x^+, x^-\right)$ as follows: 
\begin{itemize}
\item Reference Set ($\mathcal{R}$): A set of $K$ normal samples randomly selected from $\mathcal{D}$, defining the normal context.
\item Positive Sample ($x^+$): Another normal sample drawn from $\mathcal{D} \setminus \mathcal{R}$, representing an instance of in-context normality that should be identified as similar to $\mathcal{R}$.
\item Simple Negative Sample ($x_s^-$):  A normal sample drawn from a different dataset $\mathcal{D}^{\prime}$ of the same modality $\mathcal{M}$. This is designed to instill a coarse-grained, foundational discriminative ability to the model.
\item Hard Negative Sample ($x_h^-$): An anomalous sample from the source dataset $\mathcal{D}$, teaching the model to identify subtle, fine-grained deviations that define a true anomaly.
\end{itemize}
To process a triplet $\left(\mathcal{R},x^+, x^-\right)$ efficiently within a single forward pass, where $x^-$ can be either a simple or a hard negative, we adapt the input sequence from \Cref{eq:input} by introducing an additional special token, [NEG\_TOK]. Let $\mathbf{e}_{neg} \in \mathbb{R}^{N \times d_{\text{model}}}$ be the embedding of the negative sample, and $\mathbf{e}_{nt} \in \mathbb{R}^{1 \times d_{\text{model}}}$ be the embedding for [NEG\_TOK]. The full training sequence is constructed as: 
\begin{equation}
\begin{aligned}
S_{train} = {} & \text{Concat}\big( \mathbf{e}_{p}, \mathbf{e}_{ref}, \mathbf{e}_{rt}, \\
      & \mathbf{e}_{tgt},\mathbf{e}_{tt},\mathbf{e}_{neg},\mathbf{e}_{nt} \big).
\end{aligned}
\label{eq:train}
\end{equation}
Notably, we use [TGT\_TOK] as the token for the positive sample to maintain consistency with the inference phase.

After this sequence is processed by our model, we extract the final representations for the reference set ($\mathbf{h}_{\mathcal{R}}$), the positive sample ($\mathbf{h}_{x^+}$), and the negative sample ($\mathbf{h}_{x^-}$) from their respective special token positions. Finally, these representations are used to compute the loss function $\mathcal{L}$ of our CCL. Let $s(\cdot,\cdot)$ denote the cosine similarity, then $\mathcal{L}$ is defined as:
\begin{equation}
    \mathcal{L} = \max\left(s(\mathbf{h}_{\mathcal{R}}, \mathbf{h}_{x^{+}}) - s(\mathbf{h}_{\mathcal{R}}, \mathbf{h}_{x^{-}}) + \alpha,0\right),
\end{equation}
where $\alpha>0$ is a margin hyperparameter that enforces a minimum distance between positive and negative pairs. By minimizing this loss, the model is explicitly trained to produce a low discrepancy score for contextually similar pairs and a high score for dissimilar ones, directly fulfilling the discriminative learning requirement of the ICAD paradigm.

\begin{table*}[ht]
\centering
\small
\begin{tabular}{c|c||ccccc|cccc}
\toprule
\multicolumn{2}{c||}{\textbf{Modality}} & \multicolumn{5}{c|}{\multirow{2}{*}{\textbf{Task-Specific AD Methods}}} & \multicolumn{4}{c}{\multirow{2}{*}{\textbf{Universal AD Methods}}}\\
\textbf{Datasets} & \textbf{Metric} & & & & & & & & & \\
\midrule

\multicolumn{2}{c||}{Time Series} &  Anoamly. & DLinear & TimesNet & OneFitsAll & Ours &NeuTraL. & UniAD & ACR & Ours  \\
\midrule
SMD & \multirow{6}{*}{F1} & 85.68 & 79.34 & 85.94 & \underline{86.92} & \textbf{88.47} & 81.47 & \underline{84.32} & 74.38 & \textbf{88.24} \\
 MSL &  & 84.12 & 85.41 & \underline{85.78} & 82.48 & \textbf{86.52} & 79.68 & \underline{81.99} & 76.43 & \textbf{85.15} \\
 SMAP & & 71.57 & 70.39 & 72.07 & \underline{72.84} & \textbf{75.27} & 64.29 & \textbf{74.02} & 69.48 & \underline{71.95} \\
 SWAT & & 84.29 & 89.25 & 92.37 & \underline{94.27} & \textbf{94.55} & 77.43 & 79.38 & \textbf{90.70} & \underline{85.98} \\
 PSM  & & 82.36 & 93.70 & \underline{97.33} & 97.16 & \textbf{97.64} & 91.64 & \underline{92.84} & 89.53 & \textbf{96.97} \\
  \textbf{Average}  & & 81.60 & 83.62 & 86.70 & \underline{86.73} & \textbf{88.49} & 78.90 & \underline{82.51} & 80.10 & \textbf{85.66} \\
\midrule

\multicolumn{2}{c||}{Tabular} &  IForest & DAGMM & GOAD & MCM & Ours & NeuTraL. & UniAD & ACR & Ours  \\
\midrule
Cardio & \multirow{9}{*}{AUROC} & 79.67 & 66.61 & 86.27 & \underline{94.34} & \textbf{94.69} & 63.75 & 62.71 & \underline{65.51} & \textbf{91.03} \\
Campaign &  & 69.77 & 75.60 & 83.28 & \underline{87.32} & \textbf{87.44} & \underline{56.94} & 50.80 & 53.05 & \textbf{85.31} \\
Fraud  & & 73.63 & 81.47 & 78.09 & \textbf{93.23} & \underline{92.64} & 67.07 & \underline{74.38} & 65.75 & \textbf{85.26} \\
HTTP &  & 86.70 & 92.47 & 96.78 & \underline{97.77} & \textbf{98.14} & \textbf{94.18} & 88.61 & 86.75 & \underline{91.71} \\
Optdigits &  & 79.73 & 64.29 & 79.62 & \underline{97.32} & \textbf{97.88} & 63.26 & \underline{63.32} & 61.14 & \textbf{90.23} \\
Shuttle &  & 93.18 & 90.11 & 96.07 & \textbf{99.31} & \underline{98.74} & 88.14 & 90.75 & \underline{92.24} & \textbf{95.34} \\
SMTP &  & 86.73 & 88.94 & 90.08 & \underline{91.47} & \textbf{92.46} & \textbf{90.41} & 83.04 & 83.45 & \underline{86.52} \\
Wbc &  & 84.51 & 77.86 & 85.31 & \underline{97.89} & \textbf{99.06} & \underline{85.17} & 76.91 & 77.06 & \textbf{97.94} \\
\textbf{Average}  & & 81.74 & 79.67 & 86.94 & \underline{94.83} & \textbf{95.13} & \underline{76.12} & 73.82 & 73.12 & \textbf{90.42} \\ 
\midrule

\multicolumn{2}{c||}{Log} &  LogCluster & DeepLog & LogAnomaly & LogBert & Ours &  NeuTraL. & UniAD & ACR & Ours  \\
\midrule
 BGL      &   \multirow{5}{*}{AUROC}  & 83.72 & 90.29 & 82.35 & \underline{93.66} & \textbf{95.32} & 75.39 & 77.24 & \underline{84.66} & \textbf{92.79} \\
 Thunderbird &   & 74.28 & 91.88 & \underline{93.24} & 92.37 & \textbf{94.84} & 67.36 & \underline{82.31} & 78.75 & \textbf{85.20} \\
 Liberty2   &   & 83.24 & 86.27 & 93.63 & \underline{94.29} & \textbf{98.47} & 72.55 & \underline{86.29} & 84.06 & \textbf{88.69} \\
 Spirit2   &    & 88.79 & 95.25 & 92.89 & \underline{95.27} & \textbf{97.24} & 81.49 & 79.17 & \underline{87.52} & \textbf{90.44} \\
 \textbf{Average}  & & 82.51 & 90.92 & 90.53 & \underline{93.90} & \textbf{96.47} & 74.20 & 81.25 & \underline{83.75} & \textbf{89.28} \\
\bottomrule
\end{tabular}
\caption{Performance comparison of different anomaly detection methods on time series, tabular, and log data. The best and second-best results in each category are shown in bold and with an underline, respectively.}
\label{tab:anomaly_comparison}
\end{table*}

\subsection{Anomaly Detection During Inference}
During model inference, the setup is simplified. Given a reference set $\mathcal{R}$ and a test sample $x_{test}$, we
use the \method{} model to compute their representations $\mathbf{h}_{\mathcal{R}}$ and $\mathbf{h}_{x_{test}}$. The discrepancy score $\delta\left(\mathcal{R},x_{test}\right)$ is then calculated as the distance between them:
\begin{equation}
\delta\left(\mathcal{R},x_{test}\right) = \frac{1-s(\mathbf{h}_{\mathcal{R}},\mathbf{h}_{x_{test}})}{2}.
\end{equation}
This score is compared against a pre-defined threshold. If the score exceeds this threshold, the sample $x_{test}$ is classified as an anomaly relative to the context provided by $\mathcal{R}$.

\section{Experiment}
\subsection{Experimental Setup}
\subsubsection{Datasets} Our study employs a diverse collection of AD datasets spanning multiple modalities. For time series AD, we select five prominent datasets: SMD \cite{SMD}, PSM \cite{PSM}, SWaT \cite{SWaT}, MSL, and SMAP \cite{MSL}. To evaluate performance on tabular data, we incorporate 18 real-world datasets sourced from ADBench \cite{ADBench}. Furthermore, for log AD, we select four widely used datasets, including BGL, Thunderbird, Liberty2, and Spirit2 \cite{BGL}.

\subsubsection{Metrics}
Our evaluation strategy employs distinct metrics tailored to the characteristics of different data modalities. For both tabular and log datasets, we utilize AUROC as our evaluation metric. 
For time series, we follow prior works \cite{shen2020timeseries,AnomalyTrans} and employ F1-score with point adjustment. 

\subsubsection{Implementation Details}

We use Qwen2.5-0.5B \cite{qwen2.5} as the pre-trained backbone. During training, $K=5$ samples are randomly selected to form the reference set ($\mathcal{R}$), with an 8:2 ratio of simple to hard negatives. The model is trained for 5 epochs with a learning rate of 1e-5, sampling 200k instances per epoch across modalities.

\subsubsection{Baseline Methods}
To provide a comprehensive evaluation of \method{}'s performance, we compare it against two distinct categories of methods: task-specific baselines and universal baselines. 
For time series AD, we include Anomaly Transformer \cite{AnomalyTrans}, DLinear \cite{Dlinear}, TimesNet \cite{TimesNet}, and OneFitsAll \cite{OneFitsAll}; for tabular AD, we evaluate against Isolation Forest \cite{iforest}, DAGMM \cite{DAGMM}, GOAD \cite{GOAD}, and MCM \cite{MCM}; and for log AD, we compare with LogCluster \cite{LogCluster}, DeepLog \cite{DeepLog}, LogAnomaly \cite{LogAnomaly}, and LogBert \cite{LogBert}. To assess the capability of handling diverse data types and tasks within a single framework, we also compare \method{} against universal AD methods including NeuTraL AD \cite{NeutralAD}, UniAD \cite{UniAD} and ACR \cite{ACR}. They are trained across all datasets used in our experiment, mirroring the training scope of \method{} to ensure a fair comparison.

\subsection{Main Results}
\Cref{tab:anomaly_comparison} summarizes the comprehensive AD performance of \method{} against competitive baselines across multiple modalities, with results structured to highlight two key comparisons: task-specific and universal AD. Notably, \Cref{tab:anomaly_comparison} summarizes the results on 8 key tabular datasets, while the full results of all 18 datasets are provided in the appendix.

\subsubsection{Comparison with Task-Specific Methods}
We first evaluate \method{} against task-specific baselines within each modality. While each baseline model is trained individually for a single dataset, \method{} uses a single model, trained only once on a composite dataset comprising all tasks within that modality. As shown in \Cref{tab:anomaly_comparison}, \method{} consistently achieves competitive or superior performance across nearly all benchmarks (e.g., $\uparrow 4.18$ on Liberty2 compared to the second-best result). Our approach proves that even a single model can develop a sufficient understanding of anomalies compared to specialized counterparts.
\begin{figure}[t]
    \centering
    \includegraphics[width=0.95\linewidth]{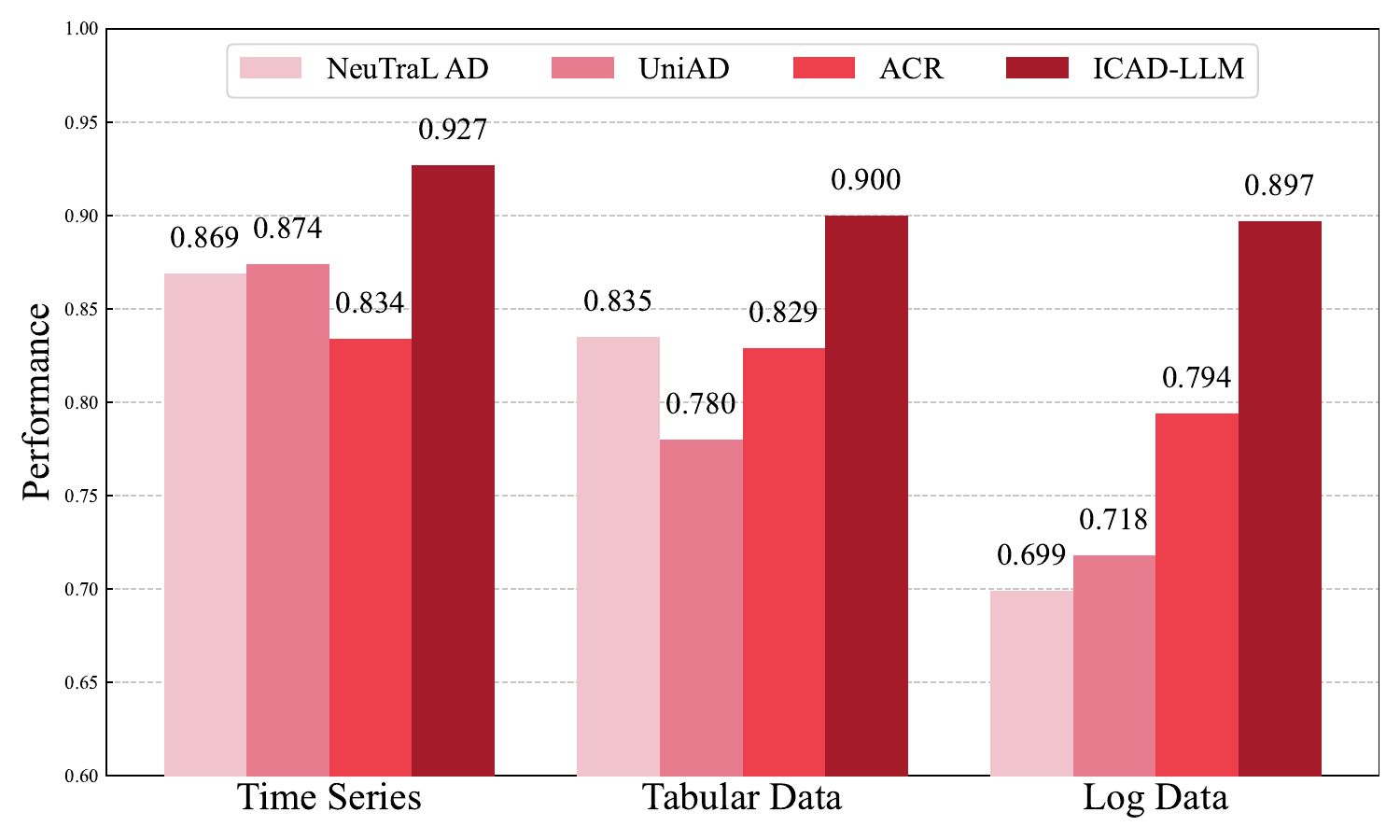}
    \caption{Generalization performance comparison different methods on unseen datasets across three modalities.}
    \label{fig:generalization}
\end{figure}

\subsubsection{Comparison with Universal Methods}
We evaluate the performance of universal anomaly detection methods trained jointly on datasets from all three modalities—time series, tabular, and log data—using a single shared model. Under this challenging setting, \method{} delivers performance that closely approaches and occasionally even surpasses strong task-specific baselines. 
In contrast, existing universal AD methods yield results that are generally inferior to task-specific approaches and often lack reliability across diverse datasets. These observations demonstrate \method{}'s strong adaptability and its practical value in meeting the growing demand for efficient, scalable anomaly detection solutions capable of handling diverse modalities within a single model.

\subsection{Generalization Experiment}
To evaluate \method{}'s ability to generalize to unseen data, we conduct experiments where the model is tested on datasets that were entirely excluded from training. This evaluation comprises one time series dataset, four tabular datasets, and one log dataset. For the tabular modality, we report the average performance. Experiment details and complete results are provided in the appendix. As shown in \Cref{fig:generalization}, \method{} maintains strong performance on these unseen datasets, outperforming all baselines across every data modality. In contrast, the baselines exhibit inconsistent performance across different data types, highlighting their limited adaptability. These results validate the robustness and effective generalization capability of our model.

\begin{figure}[t]
    \centering
    \includegraphics[width=0.9\linewidth]{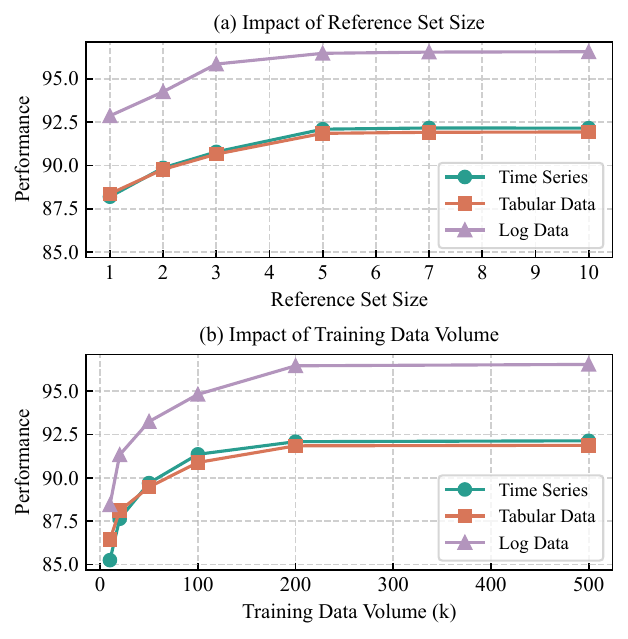}
    \caption{
    Impact of (a) reference set size and (b) total training data volume on the performance of \method{}.}
    \label{fig:sensitivity}
\end{figure}
\subsection{Sensitivity Analysis}
To thoroughly understand the contribution of key design choices in \method{}, we conduct sensitivity analysis to investigate two critical factors: the number of samples in the reference set and the total volume of training data. Detailed experimental setups and the complete results of this analysis are provided in the appendix.

\subsubsection{Impact of Reference Set Size}
We assess the influence of the reference set size, $K$, by varying it across a range of values, with a focus on smaller sizes ($K=1,2,3,5$) and also including larger values ($K=7,10$) to observe the trend. As shown in \Cref{fig:sensitivity}(a), it is observed that as $K$ increases, the average performance initially rises rapidly. However, beyond $K=5$, the performance improvement becomes notably slower. This phenomenon may be attributed to the representativeness of the reference set: smaller reference set sizes may not adequately capture the common characteristics of normal instances, while larger sizes yield diminishing marginal returns as the informative content becomes saturated. Consequently, $K=5$ is selected as the reference set size for all other experiments in our study. 

\subsubsection{Impact of Total Training Data volume}
We evaluate six different scales of total training data volumes, ranging from 10k to 500k samples, sourced from all datasets across the various modalities. As presented in \Cref{fig:sensitivity}(b), larger data volumes consistently lead to improved model performance. However, beyond 200k samples, the performance gains become marginal while the training cost increases significantly. Therefore, considering this clear performance-cost trade-off, we adopt 200k samples as the standard training volume in our work.


\section{Conclusion}
In this paper, we propose In-Context Anomaly Detection (ICAD), which reframes anomaly detection as a dynamic in-context comparison rather than memorizing a fixed normal distribution. Our model, \method{}, uses a Large Language Model to learn a general discrepancy function, enabling a single training process to handle diverse modalities and unseen tasks without retraining. Extensive experiments show that \method{} matches specialized methods and demonstrates strong generalization. This work advances the vision of a One-for-ALL AD framework, offering a viable path for developing more scalable and adaptable systems for real-world applications.

\section{Acknowledgments}
Jingyuan Wang's work was partially supported by the National Natural Science Foundation of China (No. 72171013, 72222022, 72242101) and the Fundamental Research Funds for the Central Universities (JKF-2025017226182). Juhua Pu's work was partially supported by the National Natural Science Foundation of China (No. 62577006).

\bibliography{aaai2026}

\clearpage

\appendix

\section{Appendix}

\subsection{Implementation Details}
\subsubsection{Model Architecture and Hyperparameters}
All experiments were conducted on a server equipped with 8 NVIDIA RTX 4090 GPUs. We use Qwen2.5-0.5B as the base model. The model was trained using the Adam optimizer with a learning rate of 1e-5, and a batch size of 64, 256, and 32 for time series, tabular data, and log data, respectively. We trained the model for a total of 5 epochs on the 200k combined dataset. Our implementation is based on Python 3.10, PyTorch 2.6.0, and the Transformers 4.47.1.

\subsubsection{Prompt Formulation}
To guide the LLM's contextual reasoning process, we designed a specific instruction prompt for the Prompt-Guided Representation Module. This prompt structures the task for the model, instructing it to perform a comparison rather than a simple language understanding. The prompt template used in our experiments is as follows:
\begin{quote}
    \textit{``Determine if the sample exhibits any significant discrepancies or anomalies by comparing it to the reference set.''}
\end{quote}
This instruction serves as a prefix to the concatenated embeddings of the target and reference samples, forming the final input for the LLM.

\subsubsection{Data Splitting Details}
\begin{itemize}
    \item  Task-specific Setting: For a dataset $i$ within a single modality, the sampling probability $p_i$ is proportional to its size $N_i$ (number of samples): $p_i = N_i / \sum_j N_j$. For time-series datasets, to avoid severe under-sampling of smaller sets, we adjust the effective size to $N'_i = max(N_i, 250k \text{ time points})$, preventing smaller datasets from being under-sampled.
    \item  Universal Setting: To ensure balanced training across modalities, we first select a modality (Tabular, Time-series, or Log) with uniform probability (1/3). Then, within the chosen modality, we sample triplets from its constituent datasets using the same proportional-to-size strategy as in the task-specific setting.
\end{itemize}

\subsection{Complete Results of Main Experiment}
For brevity, the main paper presents results on a representative subset of 8 tabular datasets. Table 2 provides the complete performance comparison of \method{} against baselines on all 18 tabular datasets. As shown, \method{} consistently ranks as the best or second-best method across all datasets, supporting the conclusions drawn in the main text.

\subsection{Generalization Experiment Details}
\subsubsection{Experimental Setups}
In the generalization experiment, we test the models on datasets entirely excluded from the training pool to evaluate their zero-shot adaptation capability. The selected datasets (PSM for time series; HTTP, Shuttle, SMTP, and Wbc for tabular data; BGL for log data) are widely used benchmarks representing diverse real-world scenarios. The training set composition was adjusted to maintain the total volume at 200k samples, with the reference set size $K$ fixed at 5.

\subsubsection{Detailed Results}
\Cref{tab:generalization-tab} presents the detailed scores for each method on the held-out datasets. The results show that \method{} maintains high performance across all modalities, whereas the performance of baseline methods varies significantly.

\subsection{Sensitivity Analysis Details}
\subsubsection{Impact of Reference Set Size}
This experiment investigates the model's sensitivity to the reference set size. We vary $K$ within the range of $\{1,2,3,5,7,10\}$ while keeping the total training data volume fixed at 200k. \Cref{tab:ref-size} provides the detailed results for each dataset.

\subsubsection{Impact of Training Data Volume}
This experiment analyzes the effect of the total training data volume on model performance. We vary the volume from 10k to 500k samples, with the reference set size $K$ fixed at 5. Table 5 shows the detailed results. The trend of diminishing returns observed in the main paper is evident across most individual datasets, confirming that 200k samples is a data-efficient choice.

\begin{table*}[ht]
\centering
\begin{threeparttable}[b]
\small
\begin{tabular}{c|c||ccccc|cccc}
\toprule
\multicolumn{2}{c||}{\textbf{Modality}} & \multicolumn{5}{c|}{\multirow{2}{*}{\textbf{Task-Specific AD Methods}}} & \multicolumn{4}{c}{\multirow{2}{*}{\textbf{Universal AD Methods}}}\\
\textbf{Datasets} & \textbf{Metric} & & & & & & & & & \\
\midrule

\multicolumn{2}{c||}{Time Series} &  Anoamly.$^*$ & DLinear & TimesNet & OneFitsAll & Ours &NeuTraL. & UniAD & ACR & Ours  \\
\midrule
SMD & \multirow{6}{*}{F1} & 85.68 & 79.34 & 85.94 & \underline{86.92} & \textbf{88.47} & 81.47 & \underline{84.32} & 74.38 & \textbf{88.24} \\
 MSL &  & 84.12 & 85.41 & \underline{85.78} & 82.48 & \textbf{86.52} & 79.68 & \underline{81.99} & 76.43 & \textbf{85.15} \\
 SMAP & & 71.57 & 70.39 & 72.07 & \underline{72.84} & \textbf{75.27} & 64.29 & \textbf{74.02} & 69.48 & \underline{71.95} \\
 SWAT & & 84.29 & 89.25 & 92.37 & \underline{94.27} & \textbf{94.55} & 77.43 & 79.38 & \textbf{90.70} & \underline{85.98} \\
 PSM  & & 82.36 & 93.70 & \underline{97.33} & 97.16 & \textbf{97.64} & 91.64 & \underline{92.84} & 89.53 & \textbf{96.97} \\
  \textbf{Average}  & & 81.60 & 83.62 & 86.70 & \underline{86.73} & \textbf{88.49} & 78.90 & \underline{82.51} & 80.10 & \textbf{85.66} \\
\midrule

\multicolumn{2}{c||}{Tabular} &  IForest & DAGMM & GOAD & MCM & Ours & NeuTraL. & UniAD & ACR & Ours  \\
\midrule
Breastw  & \multirow{19}{*}{AUROC} &  86.20 & 74.07 & 83.28 & \underline{99.45} & \textbf{99.67} & 67.64 & \underline{80.72} & 72.96 & \textbf{91.73} \\
Cardio &  & 79.67 & 66.61 & 86.27 & \underline{94.34} & \textbf{94.69} & 63.75 & 62.71 & \underline{65.51} & \textbf{91.03} \\
Campaign &  & 69.77 & 75.60 & 83.28 & \underline{87.32} & \textbf{87.44} & \underline{56.94} & 50.80 & 53.05 & \textbf{85.31} \\
Cardiotocography  &  &  70.53 & \underline{79.18} & 73.92 & 78.84 & \textbf{81.21} & 52.32 & \underline{67.55} & 61.39 & \textbf{74.86} \\
Fraud  & & 73.63 & 81.47 & 78.09 & \textbf{93.23} & \underline{92.64} & 67.07 & \underline{74.38} & 65.75 & \textbf{85.26} \\
Glass  &  &  64.80 & 68.57 & 65.09 & \underline{69.34} & \textbf{70.50} & 58.01 & \underline{63.96} & 52.35 & \textbf{66.95} \\
HTTP &  & 86.70 & 92.47 & 96.78 & \underline{97.77} & \textbf{98.14} & \textbf{94.18} & 88.61 & 86.75 & \underline{91.71} \\
Ionosphere  &  &  75.59 & 69.49 & 89.26 & \underline{96.91} & \textbf{99.18} & 75.75 & 66.08 & \underline{80.94} & \textbf{97.64} \\
Mammography  &  &  81.69 & 73.40 & 81.77 & \underline{89.91} & \textbf{91.93} & \underline{76.63} & 69.76 & 75.33 & \textbf{84.58} \\
Optdigits &  & 79.73 & 64.29 & 79.62 & \underline{97.32} & \textbf{97.88} & 63.26 & \underline{63.32} & 61.14 & \textbf{90.23} \\
Pima  &  &  65.36 & 59.65 & 71.80 & \underline{74.86} & \textbf{75.38} & \underline{61.16} & 58.05 & 59.38 & \textbf{68.39} \\
Pendigits  &  &  84.97 & 68.07 & 86.50 & \underline{97.33} & \textbf{99.68} & 54.29 & 58.89 & \underline{73.83} & \textbf{96.90} \\
Satellite  &  &  \underline{80.48} & 72.84 & 73.91 & 78.57 & \textbf{80.93} & 67.98 & \textbf{80.76} & 74.95 & \underline{75.13} \\
Satimage-2  &  &  91.45 & 87.54 & \underline{97.41} & 97.32 & \textbf{97.72} & 78.86 & 82.49 & \underline{89.92} & \textbf{95.35} \\
Shuttle &  & 93.18 & 90.11 & 96.07 & \textbf{99.31} & \underline{98.74} & 88.14 & 90.75 & \underline{92.24} & \textbf{95.34} \\
SMTP &  & 86.73 & 88.94 & 90.08 & \underline{91.47} & \textbf{92.46} & \textbf{90.41} & 83.04 & 83.45 & \underline{86.52} \\
Wbc &  & 84.51 & 77.86 & 85.31 & \underline{97.89} & \textbf{99.06} & \underline{85.17} & 76.91 & 77.06 & \textbf{97.94} \\
Wine  &  &  63.79 & 86.48 & 83.43 & \underline{93.88} & \textbf{96.35} & 80.54 & 87.93 & \underline{88.84} & \textbf{89.45} \\
\textbf{Average}  &  &  78.82 & 76.48 & 83.44 & \underline{90.84} & \textbf{91.87} & 71.23 & 72.60 & \underline{73.05}& \textbf{86.91} \\
\midrule
\multicolumn{2}{c||}{Log} &  LogCluster & DeepLog & LogAnomaly & LogBert & Ours &  NeuTraL. & UniAD & ACR & Ours  \\
\midrule
 BGL      &   \multirow{5}{*}{AUROC}  & 83.72 & 90.29 & 82.35 & \underline{93.66} & \textbf{95.32} & 75.39 & 77.24 & \underline{84.66} & \textbf{92.79} \\
 Thunderbird &   & 74.28 & 91.88 & \underline{93.24} & 92.37 & \textbf{94.84} & 67.36 & \underline{82.31} & 78.75 & \textbf{85.20} \\
 Liberty2   &   & 83.24 & 86.27 & 93.63 & \underline{94.29} & \textbf{98.47} & 72.55 & \underline{86.29} & 84.06 & \textbf{88.69} \\
 Spirit2   &    & 88.79 & 95.25 & 92.89 & \underline{95.27} & \textbf{97.24} & 81.49 & 79.17 & \underline{87.52} & \textbf{90.44} \\
 \textbf{Average}  & & 82.51 & 90.92 & 90.53 & \underline{93.90} & \textbf{96.47} & 74.20 & 81.25 & \underline{83.75} & \textbf{89.28} \\
\bottomrule
\end{tabular}

\begin{tablenotes}
     \item[*] We replace the joint criterion in Anomaly Transformer with reconstruction error for consistency with  other baseline methods.
\end{tablenotes}
  \caption{Performance comparison of different anomaly detection methods on time series, tabular, and log data. The best and second-best results in each category are shown in bold and with an underline, respectively.}
\end{threeparttable}
\label{tab:main_full}
\end{table*}

\begin{table*}[ht]
\centering
\small
\begin{tabular}{ccccccc}
\toprule
\textbf{Modality} & \textbf{Dataset(s)} & \textbf{Metrics} & \textbf{NeuTralAD} & \textbf{UniAD} & \textbf{ACR} & \textbf{ICAD-LLM} \\
\midrule
Time Series & PSM & F1 & 86.94 & \underline{87.44} & 83.44 & \textbf{92.65} \\
\midrule
\multirow{5}{*}{Tabular} & HTTP & \multirow{5}{*}{AUROC} & \textbf{89.89} & 76.73 & 86.35 & \underline{88.37} \\
& Shuttle & & 83.75 & 87.07 & \underline{91.37} & \textbf{91.64} \\
& SMTP & & \textbf{84.10} & 75.31 & 78.86 & \underline{81.99} \\
& Wbc & & \underline{76.21} & 72.87 & 74.99 & \textbf{97.60} \\
& \textbf{Average}  & & \underline{83.49} & 78.00 & 82.90 & \textbf{89.96} \\
\midrule
Log & BGL & AUROC & 69.90 & 71.76 & \underline{79.39} & \textbf{89.65} \\

\bottomrule
\end{tabular}
\caption{Generalization performance on unseen datasets. The model is tested on datasets entirely excluded from the training pool. The best results are in bold and the second-best are underlined.}
\label{tab:generalization-tab}
\end{table*}

\begin{table*}[ht]
\centering
\small
\setlength{\tabcolsep}{5pt}
\begin{tabular}{ccccccccc}
\toprule
\textbf{Modality} & \textbf{Datasets} &\textbf{Metrics} &\textbf{K=1} & \textbf{K=2} & \textbf{K=3} & \textbf{K=5} & \textbf{K=7} & \textbf{K=10}\\
\midrule

\multirow{6}{*}{Time Series} & SMD &\multirow{6}{*}{F1} & 87.28 & 86.96 & 86.99 & 88.47 & \textbf{88.54} & \underline{88.53}\\
 & MSL && 82.91 & 85.94 & 86.07 & 86.52 & \underline{86.65} & \textbf{86.66}\\
 & SMAP && 90.60 & 90.68 & 91.18 & 93.27 & \textbf{93.31} & \underline{93.30}\\
 & SWAT && 89.62 & 92.05 & 93.86 & \textbf{94.55} & \textbf{94.55} & \underline{94.52}\\
 & PSM && 90.53 & 93.68 & 95.81 & 97.64 & \textbf{97.75} & \underline{97.72}\\
 & \textbf{Average}  && 88.19 & 89.86 & 90.78 & 92.09 & \textbf{92.16} & \underline{92.15}\\
 \midrule
 
\multirow{19}{*}{Tabular} & Breastw &\multirow{19}{*}{AUROC} & 96.33 & 98.30 & 98.38 & \underline{99.67} & 99.65 & \textbf{99.7}\\
 & Cardio && 89.27 & 90.32 & 92.00 & \underline{94.34} & \underline{94.34} & \textbf{94.36}\\
 & Campaign && 83.65 & 85.04 & 86.11 & \underline{87.44} & \textbf{87.49} & \textbf{87.49}\\
 & Cardiotocography && 78.02 & 79.8 & \textbf{81.47} & \underline{81.21} & 81.22 & 81.20\\
 & Fraud && 91.85 & 92.49 & 92.65 & 92.64 & \textbf{92.81} & \underline{92.79}\\
 & Glass && 68.68 & 68.48 & 70.36 & 70.50 & \underline{70.66} & \textbf{70.69}\\
 & HTTP && 94.68 & 97.01 & 98.14 & 98.14 & \underline{98.18} & \textbf{98.23}\\
 & Ionosphere && 96.61 & 97.43 & 97.48 & 99.18 & \underline{99.34} & \textbf{99.36}\\
 & Mammography && 86.53 & 89.94 & 90.59 & 91.93 & \underline{92.04} & \textbf{92.08}\\
 & Optdigits && 95.83 & 95.97 & 96.07 & \underline{97.88} & 97.90 & \textbf{97.96}\\
 & Pima && 73.31 & 74.05 & 74.31 & \textbf{75.38} & \underline{75.37} & 75.36\\
 & Pendigits && 95.37 & 96.62 & 97.13 & 99.68 & \textbf{99.72} & \underline{99.71}\\
 & Satellite && 77.91 & 78.20 & 78.51 & 80.93 & \underline{80.96} & \textbf{81.00}\\
 & Satimage-2 && 94.64 & 94.48 & 96.45 & 97.72 & \underline{97.79} & \textbf{97.82}\\
 & Shuttle && 93.00 & 95.82 & 96.86 & \textbf{98.74} & \textbf{98.74} & \underline{98.73}\\
 & SMTP && 86.75 & 89.52 & 90.78 & 92.46 & \underline{92.54} & \textbf{92.55}\\
 & Wbc && 95.21 & 97.09 & 98.62 & 99.06 & \underline{99.22} & \textbf{99.29}\\
 & Wine && 92.83 & 95.25 & 95.85 & 96.35 & \underline{96.49} & \textbf{96.50}\\
 & \textbf{Average}  && 88.36 & 89.77 & 90.65 & 91.85 & \underline{91.91} & \textbf{91.93}\\
 \midrule
 
\multirow{5}{*}{Log} & BGL &\multirow{5}{*}{AUROC}& 93.21 & 93.35 & 95.27 & 95.32 & \underline{95.41} & \textbf{95.45}\\
 & Thunderbird && 91.67 & 93.62 & 93.75 & 94.84 & \underline{94.91} & \textbf{94.92}\\
 & Liberty && 95.56 & 96.44 & 98.67 & 98.47 & \textbf{98.76} & \underline{98.75}\\
 & spirit && 91.01 & 93.58 & 95.69 & \textbf{97.24} & 97.07 & \underline{97.13}\\
 & \textbf{Average}  && 92.86 & 94.25 & 95.85 & 96.47 & \underline{96.54} & \textbf{96.56}\\

\bottomrule
\end{tabular}
\caption{Sensitivity analysis with respect to the reference set size $K$. Performance is reported for each dataset as $K$ varies. The best and second-best results for each dataset are in bold and underlined, respectively.}
\label{tab:ref-size}
\end{table*}

\begin{table*}[ht]
\centering
\small
\setlength{\tabcolsep}{5pt}
\begin{tabular}{ccccccccc}
\toprule
\textbf{Modality} & \textbf{Datasets} &\textbf{Metrics} & \textbf{10K} & \textbf{20K} & \textbf{50K} & \textbf{100K} & \textbf{200K} & \textbf{500K}\\
\midrule

\multirow{6}{*}{Time Series} & SMD &\multirow{6}{*}{F1} & 81.71 & 85.63 & 85.31 & 87.39 & \underline{88.47} & \textbf{88.48}\\
& MSL && 78.56 & 79.34 & 83.81 & 85.46 & \underline{86.52} & \textbf{86.60}\\
& SMAP && 87.94 & 92.25 & 91.95 & 92.87 & \textbf{93.27} & \underline{93.25}\\
& SWAT && 85.24 & 88.34 & 92.39 & 94.27 & \underline{94.55} & \textbf{94.72}\\
& PSM && 92.78 & 92.64 & 95.04 & \underline{96.82} & \textbf{97.64} & \textbf{97.64}\\
& \textbf{Average} && 85.25 & 87.64 & 89.71 & 91.36 & \underline{92.09} & \textbf{92.14}\\
\midrule
\multirow{19}{*}{Tabular} & Breastw &\multirow{19}{*}{AUROC}& 96.19 & 96.59 & 99.55 & \underline{99.57} & \textbf{99.67} & \textbf{99.67}\\
& Cardio && 83.35 & 87.11 & 90.44 & 92.84 & \underline{94.34} & \textbf{94.54}\\
& Campaign && 77.89 & 81.91 & 84.26 & 85.98 & \underline{87.44} & \textbf{87.46}\\
& Cardiotocography && 71.64 & 75.39 & 78.66 & 81.20 & \underline{81.21} & \textbf{81.25}\\
& Fraud && 88.34 & 90.75 & 90.41 & 91.66 & \textbf{92.64} & \underline{92.62}\\
& Glass && 65.71 & 67.46 & 67.44 & 69.81 & \textbf{70.50} & \underline{70.44}\\
& HTTP && 92.71 & 93.17 & 93.98 & 95.95 & \underline{98.14} & \textbf{98.20}\\
& Ionosphere && 95.85 & 95.97 & 96.85 & 97.98 & \textbf{99.18} & \underline{99.10}\\
& Mammography && 84.85 & 87.19 & 89.62 & 91.06 & \underline{91.93} & \textbf{92.09}\\
& Optdigits && 94.06 & 95.26 & 97.15 & \textbf{98.06} & \underline{97.88} & 97.86\\
& Pima && 70.15 & 72.47 & 73.00 & 73.81 &\underline{75.38} & \textbf{75.49}\\
& Pendigits && 95.73 & 96.89 & 98.81 & 99.47 & \underline{99.68} & \textbf{99.82}\\
& Satellite && 77.85 & 79.05 & 78.91 & 79.69 & \textbf{80.93} & \underline{80.85}\\
& Satimage-2 && 94.93 & 94.53 & 95.70 & 97.47 & \underline{97.72} & \textbf{97.74}\\
& Shuttle && 93.45 & 93.82 & 95.51 & 96.64 & \textbf{98.74} & \underline{98.68}\\
& SMTP && 87.88 & 90.91 & 91.01 & 91.09 & \underline{92.46} & \textbf{92.64}\\
& Wbc && 93.28 & 93.24 & 95.42 & 97.49 & \underline{99.06} & \textbf{99.17}\\
& Wine && 92.06 & 94.03 & 93.79 & \underline{96.32} & \textbf{96.35} & \underline{96.32}\\
& \textbf{Average} && 86.44 & 88.10 & 89.47 & 90.89 & \underline{91.85} & \textbf{91.88}\\
\midrule
\multirow{5}{*}{Log} & BGL & \multirow{5}{*}{AUROC} & 84.89 & 87.06 & 90.24 & 92.83 & \underline{95.32} & \textbf{95.42}\\
& Thunderbird && 86.78 & 90.97 & 94.06 & 94.52 & \underline{94.84} & \textbf{94.96}\\
& Liberty && 91.77 & 94.55 & 94.49 & 96.92 & \underline{98.47} & \textbf{98.54}\\
& spirit && 90.30 & 92.75 & 94.21 & 95.01 & \underline{97.24} & \textbf{97.27}\\
& \textbf{Average} && 88.44 & 91.33 & 93.25 & 94.82 & \underline{96.47} & \textbf{96.55}\\
\bottomrule
\end{tabular}
\caption{Sensitivity analysis with respect to the total training data volume. Performance is reported for each dataset across different training data sizes. The best and second-best results for each dataset are in bold and underlined, respectively.}
\label{tab:training volume}
\end{table*}

\end{document}